\pdfoutput=1
\documentclass[11pt]{article}

\usepackage[final]{acl}

\usepackage{times}
\usepackage{latexsym}
\usepackage[T1]{fontenc}
\usepackage[utf8]{inputenc}
\usepackage{microtype}
\usepackage{inconsolata}
\usepackage{graphicx}
\usepackage{xcolor}
\usepackage{booktabs}
\usepackage{tabularx}
\usepackage{amsmath}
\usepackage{amssymb}
\usepackage{amsfonts}
\usepackage{graphicx}
\usepackage{multirow}
\usepackage{caption}
\usepackage{subcaption}
\usepackage{float}
\DeclareMathOperator*{\argmax}{arg\,max}
\DeclareMathOperator*{\argmin}{arg\,min}
\usepackage{algorithm}
\usepackage{algpseudocode}
\usepackage{pdfpages}

\title{LLM-based Hierarchical Concept Decomposition for Interpretable Fine-Grained Image Classification}

\author{Renyi Qu \\
  University of Pennsylvania \\
  \texttt{requ@seas.upenn.edu} \\\And
  Mark Yatskar \\
  University of Pennsylvania \\
  \texttt{myatskar@seas.upenn.edu} \\}

\begin{document}
\maketitle
\begin{abstract}
(Renyi Qu's Master's Thesis) Recent advancements in interpretable models for vision-language tasks have achieved competitive performance; however, their interpretability often suffers due to the reliance on unstructured text outputs from large language models (LLMs). This introduces randomness and compromises both transparency and reliability, which are essential for addressing safety issues in AI systems. We introduce a novel framework designed to enhance model interpretability through structured concept analysis. Our approach consists of two main components: (1) We use GPT-4 to decompose an input image into a structured hierarchy of visual concepts, thereby forming a visual concept tree. (2) We then employ an ensemble of simple linear classifiers that operate on concept-specific features derived from CLIP to perform classification. Our approach not only aligns with the performance of state-of-the-art models but also advances transparency by providing clear insights into the decision-making process and highlighting the importance of various concepts. This allows for a detailed analysis of potential failure modes and improves model compactness, therefore setting a new benchmark in interpretability without compromising the accuracy.
\end{abstract}

\section{Introduction}
Consider a scenario in wildlife conservation where researchers need a model to classify species from camera trap images. A biologist might need to understand what specific patterns the model uses to identify certain images as a threatened species - Is the inference based on the differences in fur color, body shape, or perhaps confusing background elements with the animal? Similarly, adjusting the model's input by clarifying that a visual feature is a plant rather than an animal part could potentially alter its classification. Furthermore, if new information becomes available, such as unexpected rainfall in which certain animals change their fur color, how might this change the model's predictions? The lack of clear, structured interpretability not only hinders collaboration between human experts and AI systems but also limits the ability to trust and effectively manage these tools in dynamic, real-world environments. This is especially critical in high-stakes scenarios with fine-grained inference tasks, which involve identifying the subtle differences between similar subcategories within a category \citep{safety}.

There are two main approaches to enhancing model interpretability: post-hoc explanations, which often fail to consistently reflect the model's actual decision-making processes \citep{stop}, and interpretable-by-design models, which are constrained by the inherent trade-off between interpretability and performance \citep{xai,tradeoff}. Current interpretable vision-language models leverage the unstructured and stochastic text outputs from LLMs \citep{labo, pratt, fine-grained, interpretablellm, lvlm}. While effective, the inherent randomness and lack of structure can obscure the underlying reasons for their decisions, reduce the readability, and complicate the debugging process and error analysis \citep{hallucination}. Moreover, they often lack the flexibility needed to adapt to unseen scenarios. For instance, consider an image of an airplane taken from the rear. Features like the windshield are not visible, rendering typical visual cues for subclass identification irrelevant; if the airplane is damaged, standard visual indicators might also fail, which highlights the inflexibility of current interpretable-by-design approaches in accommodating such variations.

\begin{figure*}[t]
  \includegraphics[width=\textwidth]{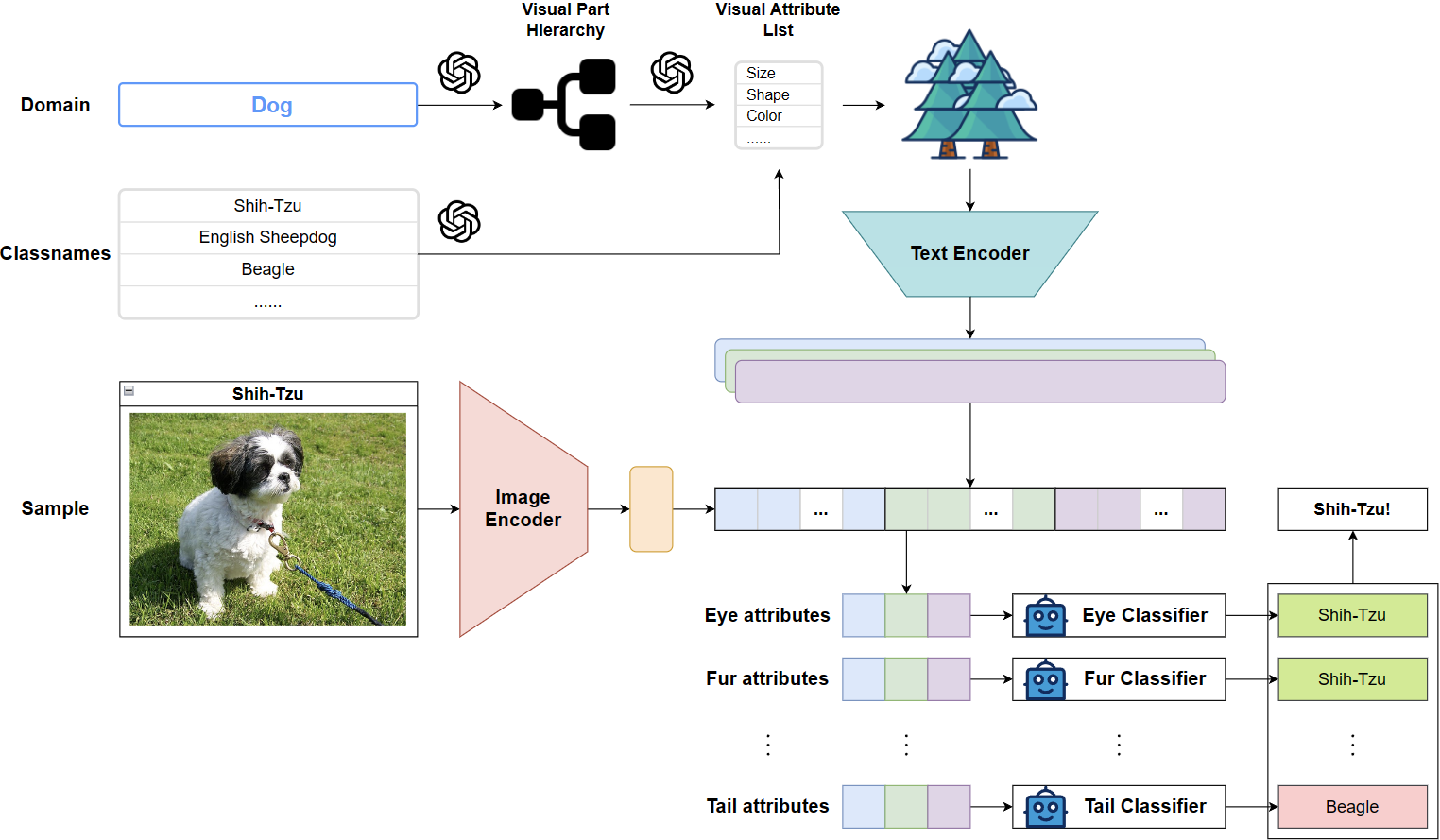}
  \caption{Our system. }
  \label{fig:experiments}
\end{figure*}

Our approach significantly advances the framework established by LaBo \citep{labo} and CBM \citep{cbm} by refining the interpretability through hierarchical concept decomposition, coupled with an ensemble of simple, linear classifiers that are straightforward to interpret and debug. Unlike traditional methods that generate random visual clues and select the optimal ones based on complex, opaque metrics, our concept generation module systematically decomposes the input object category into a clear hierarchical tree of visual parts and their associated attributes. This decomposition is consistent across different inputs of the same category, and all root-to-leaf paths in the tree are considered jointly when computing features for image classification. This comprehensive approach enables high coverage of non-standard scenarios and complex cases. Following this, each classifier within our ensemble makes decisions based on the features of a specific visual part. The final prediction is derived from a collective vote among these classifiers. This structured method not only clarifies which parts are pivotal in the classification but also elucidates why certain features are crucial in distinguishing between positive and negative examples, enhancing both transparency and reliability.

We experimented on prominent fine-grained image datasets \citep{aircraft, cub, car, dog, flower, food} and demonstrated that our model achieves performance comparable to both interpretable vision-language models and traditional SOTA image classification models. Through careful ablation studies, we determined the optimal configuration for maximizing inference accuracy involves decomposing to the level of the most detailed visual parts. Deviating from this level, either by decomposing less or more, detrimentally affects performance. In the results section, we elaborate on how our hierarchical structure enhances the model by improving readability, facilitating debugging and error analysis, and adding flexibility in feature updates. Additionally, we discuss how our model maintains compactness effectively through a pruning process that eliminates redundant or non-informative parts, further enhancing model efficiency and interpretability.

In summary, our contributions are:
\begin{enumerate}
\item We introduce fully-automated hierarchical concept decomposition using GPT-4, which generates a structured tree of visual parts and attributes, offering a level of interpretability that surpasses that of existing interpretable-by-design models.
\item We employ an ensemble of simple linear concept classifiers that enable direct inspection of each visual part and attribute, which facilitates more effective debugging and visualization while maintaining a competitive standard of classification performance.
\end{enumerate}

\section{Related Work}
\noindent\textbf{LLM-assisted vision-language systems}\hspace{0.5em} Recently, the integration of Large Language Models (LLMs) into vision-language tasks has been explored to leverage their success in language processing for improved performance and interpretability in multimodal applications. For instance, researchers have utilized LLMs as expansive knowledge bases to augment Visual Question Answering (VQA) systems by providing relevant concepts \citep{dan}. Others have implemented LLM-assisted approaches to enhance image captioning within knowledge-guided VQA models \citep{vqa_caption}. Additionally, some studies have shifted from traditional knowledge extraction to generating executable programming codes using LLMs to directly address VQA tasks \citep{vipergpt, modular_code}. Further advancements include the incorporation of LLMs into the training procedures of models like BLIP-2 and BLIVA, aiming to create multimodal LLMs specifically tailored for VQA challenges \citep{blip2, bliva}. Despite these models showing promise, they often fall short of providing clear interpretability, particularly in elucidating the reasoning processes behind their decisions. Moreover, the inherent randomness and the unstructured nature of LLMs tend to obscure the rationale for their decisions, complicating the readability, debugging, and error analysis of these systems \citep{hallucination}.

\noindent\textbf{Interpretable vision-language systems}\hspace{0.5em} Two primary strategies are employed to enhance model interpretability: post-hoc explanations and interpretable-by-design models. Post-hoc explanations are applied after model development, typically through various methods of explanation generation \citep{visual_explain, self_driving_explain, nishida, llm_explain}. However, these methods do not inherently increase the interpretability of the model itself; the core model remains a black box and these explanations often fail to accurately represent the model’s decision-making processes \citep{stop}.

In contrast, interpretable-by-design models are explicitly constructed to be understandable. A significant approach within this category is Concept Bottleneck Models (CBMs), which use high-level, human-understandable concepts as an intermediate layer. These concepts are then linearly combined to predict outcomes. For instance, one application of CBMs utilized a linear layer to integrate CLIP scores with expert-designed concepts, assessing CLIP's efficacy in concept grounding \citep{vlm_primitive}. Efforts have been made to improve the human readability of CBMs by incorporating comprehensible textual guidance \citep{cbm1, cbm2, cbm3}. As large language models (LLMs) became more prevalent, current interpretable vision-language models often rely on the stochastic and unstructured text outputs from LLMs, which introduces new challenges \citep{labo, pratt, fine-grained, interpretablellm, lvlm}. A notable drawback of CBMs is their high reliance on costly and unreliable manual annotations, which typically results in poorer performance compared to more opaque models. Attempts to mitigate these issues have included substituting the traditional knowledge base with concepts generated by LLMs, which has led to notable gains in both interpretability and performance \citep{labo}. Nevertheless, the resulting visual concepts often remain unstructured and ambiguous, limiting their effectiveness in clearly elucidating the attributes relevant to each classified image.

\section{Method}
The problem setting for fine-grained image classification is defined as follows: given an image-label pair $(i,y)$, where $i$ is a raw image and $y\in\mathcal{Y}$ is a subclass from a set of similar subclasses within the same domain $K$, we first convert the raw image into a feature representation $x=g(i)\in\mathcal{X}$. The classification model then predicts a label $\hat{y}=f(x)$. During training, the model's objective is to minimize the discrepancy between the predicted output and the actual label $\mathcal{L}(\hat{y},y)$. During inference, the prediction is used directly for evaluation or deployment purposes.

In the context of language-assisted fine-grained image classification, where an alignment model like CLIP is utilized, each sample is augmented to include a text description, forming a triplet $(i,t,y)$. This introduces an additional component to feature engineering involving text data. Image features $x_i$ are derived using an image encoder $x_i=E_I(i)\in\mathbb{R}^d$, where $d$ is the dimension of the image embeddings. Similarly, text features $x_t$ are derived using a text encoder $x_t=E_t(t)\in\mathbb{R}^d$, with the assumption that text embeddings share the same dimensional space as image embeddings. The final features $x$ are formed by integrating both image and text embeddings, $x=g(x_i,x_t)$. This enriched feature set is used to predict the subclass $\hat{y}=f(x)$.

Our method aims to enhance both interpretability and performance by improving the quality of text features $x_t$ and refining the modeling component $f(x)$. This approach encompasses two main elements: concept tree decomposition and concept classification. To clarify the discussion, we define three key terms that are used extensively throughout this section.\vspace{1em}

\noindent\textbf{Visual part}: This refers to a discernible physical segment of an object visible to the human eye. For example, within the category of dogs, 'head' is a visual part, 'mouth' is a part of the head, and 'tongue' is a part of the mouth. This is denoted as $p\in\mathcal{P}$, where $\mathcal{P}$ denotes the set of all possible visual parts for the specific class domain.\vspace{1em}

\noindent\textbf{Visual attribute}: These are observable characteristics of a visual part, such as size, shape, color, material, finish, and design complexity. For instance, visual attributes of a car’s front bumper might include its color and material. This is denoted as $a\in\mathcal{A}_p$, where $\mathcal{A}_p$ denotes the set of all possible visual attributes for a specific visual part $p$.\vspace{1em}

\noindent\textbf{Attribute value}: This is the specific manifestation of a visual attribute. For example, the color attribute of an American crow’s primary features would be black, and the shape attribute of its eyes would be rounded. This is denoted as $v\in\mathcal{V}_a$, where $\mathcal{V}$ denotes the set of all possible visual attributes for a specific visual attribute $a$.

We assume all subclasses share the same visual parts and visual attributes and differ in attribute values. Therefore, all subclasses share the same $\mathcal{P}$ and the same set $\mathcal{A}_p$ for each $p\in\mathcal{P}$.

\subsection{Concept Tree Decomposition}
Our approach decomposes an arbitrary object category into a concept tree where each intermediate node corresponds to a visual part of the category in question, each leaf node represents a visual attribute value pertinent to its parent visual attribute, and each visual attribute node bridges its parent visual part and its corresponding child value. The decomposition process unfolds in three sequential steps:

\noindent\textbf{Visual Part Decomposition}\hspace{0.5em} Starting with a given object domain $K$, we use GPT-4 to generate a hierarchical arrangement of all possible visual parts of the object via zero-shot prompting, denoted as $\mathcal{P}=\text{LLM}_\text{zero}(K)$. This hierarchy is then formatted and preserved in a JSON structure, where each node denotes a visual part identified by GPT-4.\vspace{1em}

\noindent\textbf{Visual Attribute Generation}\hspace{0.5em} For every visual part identified $p\in\mathcal{P}$, we use GPT-4 to enumerate 3-7 pertinent visual attributes via few-shot prompting, denoted as $\mathcal{A}_p=\text{LLM}_\text{few}(p,\mathcal{M}_{p\rightarrow a})$, where $\mathcal{M}_{p\rightarrow a}$ consists of three fixed examples of part-to-attribute mappings for consistency and diversity. This stage not only identifies common attributes such as size, shape, and color but also specific ones like luminosity, opacity, and thickness. The aim is to capture a diverse array of attributes, enhancing the richness of the concept tree.\vspace{1em}

\noindent\textbf{Concept Tree Generation}\hspace{0.5em} In this final step, we assign attribute values to each visual attribute $a\in\mathcal{A}_p$ of each visual part $p\in\mathcal{P}$ of each specific subclass $y\in\mathcal{Y}$, denoted as $\mathcal{V}_a^y=\text{LLM}_\text{crit}(a,p,y)$. Using self-critique prompting that involves a sequence of three critical queries, we refine these assignments:
\begin{enumerate}
\item \textbf{Logical Relationships}: We determine whether multiple attribute values should be combined using logical operators (\texttt{AND}/\texttt{OR}). For instance, if a cat's fur is described as "black \texttt{AND} white," the attribute value implies the presence of both colors simultaneously. However, if it is described as "black \texttt{OR} white", then either color satisfies this condition, so this phrase should be separated into two leaf nodes.
\item \textbf{Attribute Value Consistency}: To ensure compatibility with the CLIP text encoder, which differs in handling various Part-Of-Speech (POS) tags, we standardize multi-word and noun attribute values into an "of" attributive format, while single-word adjectives remain unchanged.
\item \textbf{Redundancy Reduction}: We scrutinize the attribute list to eliminate any repetitions that might bias the model, ensuring that each visual attribute and its values are uniquely represented.
\end{enumerate}

\begin{figure*}[h]
  \includegraphics[width=\textwidth]{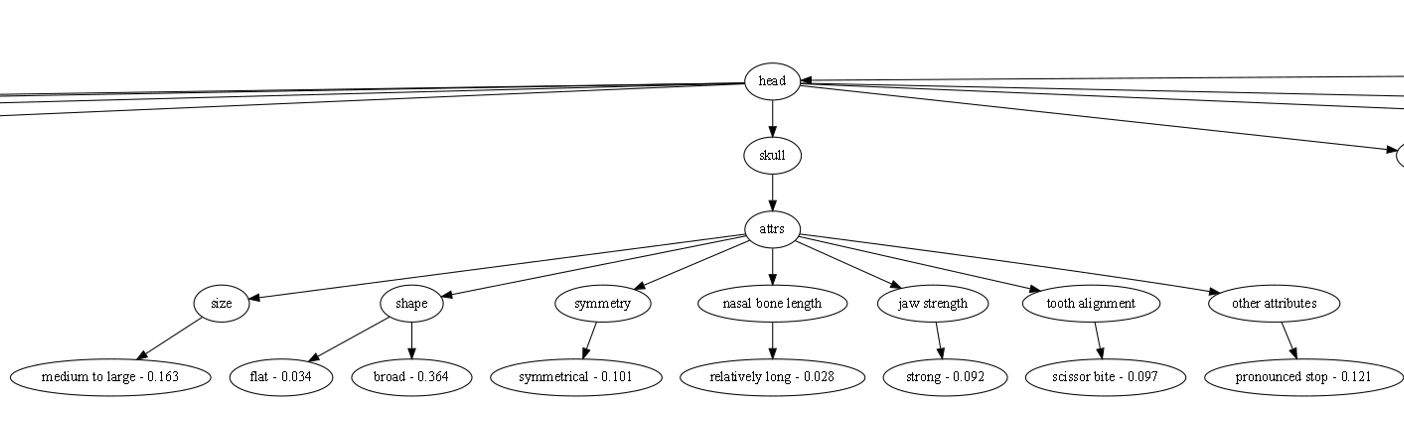}
  \caption{A snippet of a Part-Attribute subtree for the "Skull" part. The "Head" node serves as the parent visual part of "Skull". The "Attrs" node displays a comprehensive subtree of Name-Value attribute pairs associated with the "skull" part. Each score represents the contribution of its corresponding attribute value to the prediction of the "Head" classifier.}
  \label{fig:snippet1}
\end{figure*}

\begin{figure*}[h]
  \includegraphics[width=\textwidth]{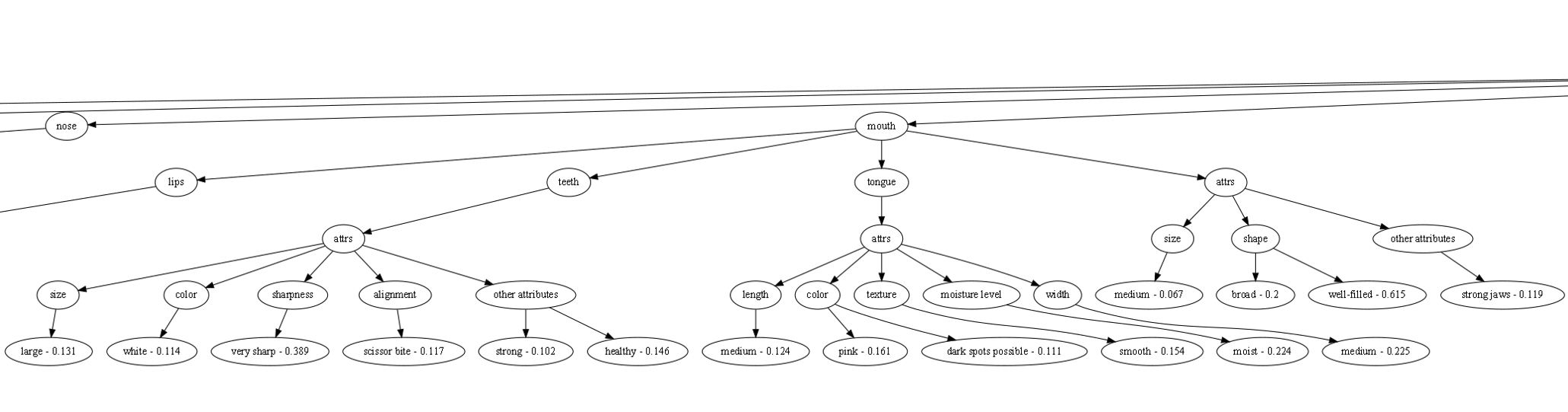}
  \caption{A snippet of the hierarchy of visual parts. The "Mouth" node originates from the "Head" parent node. Subparts of "Mouth," including "Lips," "Teeth," and "Tongue," are represented as child nodes. Each part features its own attribute subtree, independent of its level of decomposition.}
  \label{fig:snippet2}
\end{figure*}

Upon completing these steps, we convert each root-to-leaf path into a coherent natural language description denoted as $h(y,p,a,v)\in\mathcal{C}_K$, where $\mathcal{C}_K\subset\mathbb{C}$ is the subset of all visual clues for the problem domain generated by GPT-4. Note that it is infeasible to obtain the full visual clue set $\mathbb{C}$. We then encode each visual clue via the CLIP text encoder $E_T$, and store the embeddings $x_t=E_T(h(y,p,a,v))$ for subsequent classification tasks. Although trees within the same object category share structural similarities, they are distinguished by their unique attribute values. Figures \ref{fig:snippet1} and \ref{fig:snippet2} present snippets from the visual representation of the generated concept trees using Graphviz \citep{graphviz}. Due to the extensive breadth of the entire tree, it is impractical to display it in its entirety within this thesis.

\vspace{1em}

\section{Concept Classification}
Given an input image $i$, we first encode it using the CLIP image encoder $E_I$. We then calculate the similarity scores between this encoded image and each pre-computed visual clue embedding. The final features for classification $x$ are derived from the similarity function $g(x_i,x_t)=\text{sim}\left(E_I(i), E_T\left(h(y,p,a,v)\right)\right)$ across all visual clues in the set $\mathcal{C}_K$ for each image sample $i$.

Although this process could be performed by any classifier, a linear probe $\text{LP}$ is commonly utilized due to its simplicity and effectiveness. However, to enhance interpretability, we employ an ensemble of linear probes where each $\text{LP}_p$ is tasked with predicting the actual subclass $y$ based on part-specific features $x_p$. For instance, in the "dog" domain, an "eye classifier" predicts the dog breed using only the eye features, while an "ear" classifier does so using only the ear features. The part-specific features $x_p$ are extracted from the flattened feature vector $x$ by segmenting on visual parts $p$ instead of subclasses $y$, as illustrated in Figure \ref{fig:model}.

During training, each $\text{LP}_p$ is independently trained to minimize the cross-entropy loss $\mathcal{L}(\hat{y}_p, y)$. In the inference phase, each $\text{LP}_p$ generates a subclass prediction $\hat{y}_p$ and a corresponding probability vector $\textbf{P}_p=[P_p(y=1), \cdots, P_p(y=N)]$, where $P$ denotes probability and $N$ is the total number of subclasses. The final subclass prediction $\hat{y}$ is then determined either by a majority vote $\hat{y}=\text{mode}(\hat{\textbf{y}})$ or by a top-probability vote $\hat{y}=\argmax_{j}\sum_{p\in\mathcal{P}}P_p(y=j)$.

In summary, our method follows the following pipeline:
\begin{algorithm}[h]
\caption{Hierarchical Concept Decomposition}\label{alg:model}
\begin{algorithmic}
\Require $K,\mathcal{Y}, \mathcal{M}_{p\rightarrow a}$
% \Ensure $y = x^n$
\State $\mathcal{C}_K = \{\}$
\State $\mathcal{P}\gets\text{LLM}_\text{zero}(K)$ \Comment{Part Decomposition}
\For{$p\in\mathcal{P}$}
    \Comment{Attribute Generation}
    \State $\mathcal{A}_p\gets\text{LLM}_\text{few}(p,\mathcal{M}_{p\rightarrow a})$
    \For{$y\in\mathcal{Y}$}
        \For{$a\in\mathcal{A}_p$} \Comment{Tree Generation}
            \State $\mathcal{V}_a^y\gets\text{LLM}_\text{crit}(a,p,y)$
            \For{$v\in\mathcal{V}_a^y$}
                \State $\mathcal{C}_K$.insert($h(y,p,a,v)$)
            \EndFor
        \EndFor
    \EndFor
\EndFor
\State 
\end{algorithmic}
\end{algorithm}

\begin{algorithm}[h]
\caption{Concept Classification}\label{alg:clf}
\begin{algorithmic}
\Require $(i,y),\mathcal{C}_K$
\State $\text{LPs}=\text{map}(p\rightarrow\text{LP})$
\State $X_t\gets E_T(\mathcal{C}_K)$ 
\State $x_i\gets E_I(i)$ 
\State $X\gets\text{sim}(x_i, X_t)$
\State $\hat{y}\gets\text{LP}(X)$
\If{training}
    \For{$p\in\mathcal{P}$}
        \State $\text{LPs}[p]\gets\argmin_\text{LP}\mathcal{L}(\text{LP}(X_p),y)$
    \EndFor
\Else
    \State Return $\text{vote}([\text{LPs}[p](X)\text{ for }p\in\mathcal{P}])$
\EndIf
\end{algorithmic}
\end{algorithm}

\section{Experiment}
\subsection{Datasets}
Our research focuses on fine-grained image classification. To this end, we selected six open-source, certified fine-grained datasets from diverse domains. Due to budget constraints associated with generating concept trees via GPT-4, we limited our study to 20 subclasses from each dataset, utilizing all available training and test images while adhering to the original train-test split ratios. The datasets we used include:
FGVC-Aircraft \citep{aircraft}, CUB-200-2011 \citep{cub}, Stanford Cars \citep{car}, Stanford Dogs \citep{dog}, Flower-102 \citep{flower}, and Food-101 \citep{food}.

\subsection{Baselines}
Our system extends the LaBo framework \citep{labo}, which serves as our primary baseline together with its baseline model, CLIP + LP. We employ several deployment strategies for LP, described below using our notations:\vspace{1em}

\noindent\textbf{CLIP+LP (image only)}: The most common approach applies LP directly to the image embeddings generated by the CLIP image encoder, particularly the Vision Transformer \citep{vit}. In our notation, the final prediction for an image is given by $\hat{y}=\text{LP}(E_I(i))$. This method remains the benchmark baseline, as no interpretable model has surpassed it due to the intrinsic trade-off between interpretability and performance \citep{xai,tradeoff}.\vspace{1em}

\noindent\textbf{CLIP+LP (image+label)}: In our work, which incorporates text clues, LP is applied to the similarity scores between image embeddings and label embeddings. These label embeddings are derived by encoding the class names directly with the CLIP text encoder. The final prediction for an image in our notation is $\hat{y}=\text{LP}(\text{sim}(E_I(i), E_T(h(y))))$, where $h(y)$ translates the label $y$ into a natural language format, following the prompting style used in CLIP's pretraining: $h(y)=$"A photo of $y$".\vspace{1em}

\noindent\textbf{LaBo}: Our approach differentiates from LaBo in the selection of the concept space $\mathcal{C}_K$ and the predictive model $f(\cdot)$. LaBo uses randomly generated visual clues from GPT-3 and a submodular optimization module to select them. It employs a learnable class-concept matrix to assign weights to each visual clue and then applies a single linear probe to the weighted combination of image-clue similarity scores. See \citet{labo} for detailed mathematical descriptions. Due to a lack of visual clues for the Stanford Car and Stanford Dog datasets from the original authors, LaBo will be excluded from these two datasets.\vspace{1em}

\begin{table*}[t]
  \centering
  \begin{tabular}{l|cccccc}
    \hline
     & \textbf{Aircraft} & \textbf{CUB} & \textbf{Car} & \textbf{Dog} & \textbf{Flower} & \textbf{Food} \\
     \hline
    SOTA (for reference) & 0.954 & 0.931 & 0.971 & 0.973 & 0.998 & 0.986 \\
    \hline
    CLIP+LP (image only)   & \textbf{0.798}         & \textbf{0.882}     & \textbf{0.973} & 0.906 & \textbf{0.985} & 0.953 \\
    CLIP+LP (image+label)   & 0.708         & 0.856     & 0.952 & 0.907 & 0.952 & 0.961   \\
    LaBo                & 0.736         & 0.819     & -     & -     & 0.955 & 0.931 \\
    Concept Classifier (ours) & 0.783 & 0.871      & 0.974         & \textbf{0.927} & 0.976  & \textbf{0.966}   \\
    \hline
  \end{tabular}
  \caption{Test Accuracy}
  \label{tab:main}
\end{table*}

\noindent\textbf{SOTA}: Given the early stage of research in interpretable vision-language models, we include state-of-the-art (SOTA) fine-grained image classification scores as benchmarks for each dataset. These are for reference only and are not intended for direct comparison, as these scores are derived from experiments encompassing all classes, not just our subsets.

\section{Evaluation}
As indicated in Table \ref{tab:main}, CLIP+LP (image only) performs well across most datasets. However, among the interpretable models, our model stands out, achieving significantly better performance than both CLIP+LP (image+label) and LaBo. Notably, our model even surpasses the CLIP+LP (image only) on the Stanford Dogs and Food-101 datasets. This clearly demonstrates that our model not only leads in performance among interpretable vision-language models in fine-grained image classification but also competes effectively with traditional, non-interpretable image classification baseline.

Although our results lag behind the SOTA benchmarks achieved using specially trained, complex neural networks with large architectures, our approach utilizes only an ensemble of linear probes and a single, pretrained CLIP model. This highlights the efficiency and potential of our simpler, more interpretable model in achieving competitive performance.

\section{Ablation Studies}
\subsection{Comparison 1: Decomposition Depth}
\begin{table*}[h]
  \centering
  \begin{tabular}{l|cccccc}
    \hline
     & \textbf{Aircraft} & \textbf{CUB} & \textbf{Car} & \textbf{Dog} & \textbf{Flower} & \textbf{Food} \\
    \hline
    LP on top-parts         & 0.722         & 0.785         & 0.823         & 0.857         & 0.944 & 0.959    \\
    LP on all-parts         & 0.756         & 0.836         & 0.873         & 0.898         & 0.948 & \textbf{0.959}   \\
    LP on Attrs             & \textbf{0.759}         & \textbf{0.841}         & \textbf{0.913}         & \textbf{0.912}         & \textbf{0.956} & 0.955   \\
    LP on Attr-vals         & 0.754         & 0.831         & 0.912         & 0.902         & 0.950 & 0.952   \\
    \hline
  \end{tabular}
  \caption{Comparison 1: Decomposition Depth}
  \label{tab:comp1}
\end{table*}
As outlined in Section 3, the two lowest levels of our hierarchy are dedicated to visual parts, visual attributes, and attribute values, respectively. When calculating similarity for a given image, each leaf node within this hierarchy is assigned a similarity score. Each visual attribute node receives a score that reflects the maximum score among all its associated leaf nodes, embodying an "OR" logic. For instance, if the "Color" attribute includes the leaf nodes "Blue" and "Red", and "Blue" scores 0.99 while "Red" scores only 0.01, this suggests that the visual part in question is predominantly blue, thus the leaf node for "Blue" is activated. Averaging the scores in this context would not be appropriate; instead, taking the maximum score is more logical.

Conversely, each visual part node is assigned an average score calculated across all its child nodes, as these nodes collectively represent the visual part. For example, the visual part representing a dog's eye might include attributes like size, shape, color, brightness, and pupil size, all considered simultaneously. Similarly, a dog's head encompasses the parts of eyes, ears, nose, and mouth, making an average score a rational choice here.

To summarize, LP on attribute values utilizes raw similarity scores for each value. LP on attributes employs the maximum score from the leaf nodes under each attribute. LP on all parts calculates an average score across all attribute nodes within each visual part node. LP on top parts averages the scores across all subpart nodes of each major visual part node located just below the root node in the hierarchy.

For this ablation study, a single linear probe was used instead of an ensemble. As demonstrated in Table \ref{tab:comp1}, attribute values are crucial for performance across most datasets, with performance increasing as the decomposition depth increases. However, delving deeper into individual attribute values can reduce performance, as some values may not be activated and thus contribute little to the final prediction, as illustrated in the "Blue" or "Red" example.

\subsection{Comparison 2: Label Inclusion}
\begin{table*}[h]
  \centering
  \begin{tabular}{l|cccccc}
    \hline
     & \textbf{Aircraft} & \textbf{CUB} & \textbf{Car} & \textbf{Dog} & \textbf{Flower} & \textbf{Food} \\
    \hline
    LP (w/o)  & 0.720         & 0.815         & 0.893         & 0.829         & 0.903 & 0.938   \\
    LP (common)         & 0.734         & 0.833         & 0.901         & 0.888         & 0.939 & 0.947   \\
    LP (with)  & \textbf{0.759}         & \textbf{0.841}         & \textbf{0.913}         & \textbf{0.912}         & \textbf{0.956} & \textbf{0.955}  \\
    \hline
    Concept Classifier (w/o)  & \textbf{0.783} & 0.871      & \textbf{0.974}         & \textbf{0.927} & \textbf{0.976}  & 0.966   \\
    Concept Classifier (common) & 0.782 & 0.870         & 0.972         & 0.926  & 0.974 & \textbf{0.968}    \\
    Concept Classifier (with)  & 0.783 & \textbf{0.871}      & 0.973         & 0.927 & 0.975 & 0.966   \\
    \hline
  \end{tabular}
  \caption{Comparison 2: Label Inclusion}
  \label{tab:comp2}
\end{table*}

\begin{table*}[h]
  \centering
  \begin{tabular}{l|cccccc}
    \hline
     & \textbf{Aircraft} & \textbf{CUB} & \textbf{Car} & \textbf{Dog} & \textbf{Flower} & \textbf{Food} \\
    \hline
    Concept Classifier (weighted)   & 0.518         & 0.529        & 0.699         & 0.638         & 0.787 & 0.754   \\
    Concept Classifier (maj)   & 0.765         & 0.841        & 0.938         & 0.913         & 0.959 & 0.962   \\
    Concept Classifier (prob)  & \textbf{0.783} & \textbf{0.871}      & \textbf{0.974}         & \textbf{0.927} & \textbf{0.976}  & \textbf{0.966}   \\
    \hline
  \end{tabular}
  \caption{Comparison 3: Voting Strategy}
  \label{tab:comp3}
\end{table*}
It is well-documented that including the label in the text prompt for the CLIP text encoder can significantly enhance its performance, as it is pretrained on texts that typically include the label \citep{clip,clipsucks,labo}. In light of this, we evaluate whether our concept classifier can effectively counter this issue. The term "w/o" denotes scenarios where the label is not included, using the format $h=$"A photo of $h(p,a,v)$". The term "with" indicates scenarios where the label is included, formatted as $h=$"A photo of $y$ with $h(p,a,v)$". "Common" refers to the inclusion of the domain name for all labels, where $h=$"A photo of $K$ with $h(p,a,v)$".

As demonstrated in Table \ref{tab:comp2}, LPs are affected by this issue, showing that the inclusion of the label name significantly outperforms the other two approaches. However, this discrepancy is mitigated when we deploy our ensemble of LPs. The variations in the prompt do not significantly impact our performance, indicating that our system maintains stability regardless of how the prompt is structured.

\subsection{Comparison 3: Voting Strategy}
In addition to the two voting strategies outlined in Section 3.2, we explored the integration of a learnable part-attribute weight matrix, analogous to the class-concept weight matrix used in the LaBo framework. Contrary to expectations, incorporating learnable weights significantly impaired our performance. Upon a detailed examination of the feature importance assigned to each attribute within each visual part classifier, we found that the feature importance was nearly uniform across all attributes. This uniformity rendered the additional training of a weight matrix unnecessary.

Furthermore, our analysis showed that the top-probability voting strategy outperforms the majority voting strategy. This suggests that while some visual part classifiers may incorrectly predict the label, they often still assign a high probability to the correct label, albeit not the highest compared to the incorrect one. This finding highlights the effectiveness of using probability-based decision metrics over simple majority rule in enhancing classification accuracy.

\section{Conclusion}
In this paper, we have presented a novel approach to fine-grained image classification that significantly enhances interpretability without sacrificing performance. Our method builds upon existing frameworks by introducing a structured hierarchical concept decomposition, coupled with an ensemble of linear classifiers, to clarify decision-making processes in vision-language models. This approach not only advances the state of interpretable-by-design models but also addresses the critical need for reliability and transparency in applications such as wildlife conservation, where understanding the nuances of model decisions is paramount. Our experiments across various fine-grained image datasets demonstrate that our model competes effectively with both current interpretable models and traditional state-of-the-art image classification systems. The integration of a hierarchical concept tree with an ensemble of classifiers ensures detailed and comprehensible insights into the classification process, thereby allowing users to pinpoint the specific visual parts and attributes influencing the model's predictions. This granularity facilitates robust debugging, precise error analysis, and the ability to adapt to new or changing scenarios, which are often challenging for more rigid models. Moreover, the simplicity of the linear classifiers within our ensemble underscores our commitment to maintaining an interpretable system that can be easily understood and manipulated by practitioners, thereby bridging the gap between complex machine learning models and practical applications that require user trust and understanding. Further research in developing a comprehensive interpretability evaluation framework is necessary to justify our claim and further facilitate the advancement of interpretable vision-language models.

\section*{Limitation}
Our approach utilizes the pretrained CLIP model without fine-tuning, which may result in suboptimal performance and not fully leverage the potential of the alignment model structure. Additionally, due to limited computational resources, we were unable to explore more advanced alignment models like BLIP. Moreover, our experiments were conducted on only six domain-specific datasets, which might not provide sufficient evidence to generalize our findings widely. We believe that these modifications could significantly enhance performance, and further experimentation is needed to substantiate this assumption.

Interpretability in vision-language modeling remains a relatively uncharted field, lacking well-established benchmarks or comprehensive evaluation frameworks to validate our claims of interpretability. Although we have developed a demonstration tool, showcased in Appendix A and soon to be made public on Github, a formally recognized and rigorously tested evaluation framework is still needed to support our claims definitively.

Throughout the project, we made several structural changes to improve the efficiency of storing and utilizing the generated concept trees, transitioning from Python dictionaries to JSON trees and finally to Pandas dataframes. The current system, based on Pandas dataframes, offers the fastest performance and greatest interpretability among the structures we tested. However, we believe that there are additional opportunities to enhance efficiency and performance further.

Due to the significant amount of time used on exploration and narrowing down the research topics, we could not run human evaluation studies, which are critical for validating the interpretability of our model. Future work will prioritize the establishment of human evaluation protocols, dividing them into two categories: concept generation and concept classification.

In concept generation, we plan to assess the generated concepts on a Likert scale for factuality, sufficiency, and compactness. Factuality will measure the accuracy of the concept hierarchy relative to the specific subclass. Sufficiency will evaluate whether the hierarchy comprehensively represents all visual aspects of the subclass. Compactness will scrutinize the hierarchy for any redundancies, suggesting improvements where necessary.

In concept classification, our focus will shift to groundability and consistency through pairwise comparisons. Groundability will examine whether the top-$k$ clues provided by our classifiers accurately and representatively describe the image compared to a baseline model (LaBo). Consistency will assess whether these clues maintain their reliability across visually similar images of the same subclass, taken from nearly identical angles.

However, the challenge of defining and measuring interpretability remains. The subjective nature of interpretability necessitates a robust, theoretically sound, and empirically backed framework. This ambitious task, suitable for an extended research endeavor such as a PhD project, involves developing a comprehensive interpretability evaluation framework. As such, while this initial study lays the groundwork, future research will be required to advance these efforts and further refine our system.

\section*{Acknowledgments}
This project is the Master's Thesis of Renyi Qu under the supervision of Prof. Mark Yatskar at the University of Pennsylvania.
\bibliography{custom}

\begin{thebibliography}{35}
\providecommand{\natexlab}[1]{#1}

\bibitem[{Bhalla(2022)}]{vlm_primitive}
Usha Bhalla. 2022.
\newblock Do vision-language pretrained models learn primitive concepts?

\bibitem[{Bossard et~al.(2014)Bossard, Guillaumin, and Van~Gool}]{food}
Lukas Bossard, Matthieu Guillaumin, and Luc Van~Gool. 2014.
\newblock Food-101 -- mining discriminative components with random forests.
\newblock In \emph{European Conference on Computer Vision}.

\bibitem[{Bujwid and Sullivan(2021)}]{cbm1}
Sebastian Bujwid and Josephine Sullivan. 2021.
\newblock Large-scale zero-shot image classification from rich and diverse textual descriptions.
\newblock \emph{arXiv preprint arXiv:2103.09669}.

\bibitem[{Chen et~al.(2023)Chen, Chen, Diao, Wan, and Wang}]{clipsucks}
Zhihong Chen, Guiming Chen, Shizhe Diao, Xiang Wan, and Benyou Wang. 2023.
\newblock On the difference of bert-style and clip-style text encoders.
\newblock In \emph{Findings of the Association for Computational Linguistics: ACL 2023}, pages 13710--13721.

\bibitem[{Dosovitskiy et~al.(2020)Dosovitskiy, Beyer, Kolesnikov, Weissenborn, Zhai, Unterthiner, Dehghani, Minderer, Heigold, Gelly et~al.}]{vit}
Alexey Dosovitskiy, Lucas Beyer, Alexander Kolesnikov, Dirk Weissenborn, Xiaohua Zhai, Thomas Unterthiner, Mostafa Dehghani, Matthias Minderer, Georg Heigold, Sylvain Gelly, et~al. 2020.
\newblock An image is worth 16x16 words: Transformers for image recognition at scale.
\newblock \emph{arXiv preprint arXiv:2010.11929}.

\bibitem[{Du et~al.(2023)Du, Li, Tang, Zhao, and Wen}]{vqa_caption}
Yifan Du, Junyi Li, Tianyi Tang, Wayne~Xin Zhao, and Ji-Rong Wen. 2023.
\newblock Zero-shot visual question answering with language model feedback.
\newblock \emph{arXiv preprint arXiv:2305.17006}.

\bibitem[{Ellson et~al.(2002)Ellson, Gansner, Koutsofios, North, and Woodhull}]{graphviz}
John Ellson, Emden Gansner, Lefteris Koutsofios, Stephen~C North, and Gordon Woodhull. 2002.
\newblock Graphviz—open source graph drawing tools.
\newblock In \emph{Graph Drawing: 9th International Symposium, GD 2001 Vienna, Austria, September 23--26, 2001 Revised Papers 9}, pages 483--484. Springer.

\bibitem[{Fu et~al.(2023)Fu, Zhang, Kwon, Perera, Zhu, Zhang, Li, Wang, Wang, Castelli et~al.}]{dan}
Xingyu Fu, Sheng Zhang, Gukyeong Kwon, Pramuditha Perera, Henghui Zhu, Yuhao Zhang, Alexander~Hanbo Li, William~Yang Wang, Zhiguo Wang, Vittorio Castelli, et~al. 2023.
\newblock Generate then select: Open-ended visual question answering guided by world knowledge.
\newblock \emph{arXiv preprint arXiv:2305.18842}.

\bibitem[{Gosiewska et~al.(2021)Gosiewska, Kozak, and Biecek}]{tradeoff}
Alicja Gosiewska, Anna Kozak, and Przemys{\l}aw Biecek. 2021.
\newblock Simpler is better: Lifting interpretability-performance trade-off via automated feature engineering.
\newblock \emph{Decision Support Systems}, 150:113556.

\bibitem[{Gunning and Aha(2019)}]{xai}
David Gunning and David Aha. 2019.
\newblock Darpa’s explainable artificial intelligence (xai) program.
\newblock \emph{AI magazine}, 40(2):44--58.

\bibitem[{Hendricks et~al.(2016)Hendricks, Akata, Rohrbach, Donahue, Schiele, and Darrell}]{visual_explain}
Lisa~Anne Hendricks, Zeynep Akata, Marcus Rohrbach, Jeff Donahue, Bernt Schiele, and Trevor Darrell. 2016.
\newblock Generating visual explanations.
\newblock In \emph{Computer Vision--ECCV 2016: 14th European Conference, Amsterdam, The Netherlands, October 11--14, 2016, Proceedings, Part IV 14}, pages 3--19. Springer.

\bibitem[{Hu et~al.(2023)Hu, Xu, Li, Li, Chen, and Tu}]{bliva}
Wenbo Hu, Yifan Xu, Y~Li, W~Li, Z~Chen, and Z~Tu. 2023.
\newblock Bliva: A simple multimodal llm for better handling of text-rich visual questions.
\newblock \emph{arXiv preprint arXiv:2308.09936}.

\bibitem[{Khosla et~al.(2011)Khosla, Jayadevaprakash, Yao, and Fei-Fei}]{dog}
Aditya Khosla, Nityananda Jayadevaprakash, Bangpeng Yao, and Li~Fei-Fei. 2011.
\newblock Novel dataset for fine-grained image categorization.
\newblock In \emph{First Workshop on Fine-Grained Visual Categorization, IEEE Conference on Computer Vision and Pattern Recognition}, Colorado Springs, CO.

\bibitem[{Kim et~al.(2018)Kim, Rohrbach, Darrell, Canny, and Akata}]{self_driving_explain}
Jinkyu Kim, Anna Rohrbach, Trevor Darrell, John Canny, and Zeynep Akata. 2018.
\newblock Textual explanations for self-driving vehicles.
\newblock In \emph{Proceedings of the European conference on computer vision (ECCV)}, pages 563--578.

\bibitem[{Koh et~al.(2020)Koh, Nguyen, Tang, Mussmann, Pierson, Kim, and Liang}]{cbm}
Pang~Wei Koh, Thao Nguyen, Yew~Siang Tang, Stephen Mussmann, Emma Pierson, Been Kim, and Percy Liang. 2020.
\newblock Concept bottleneck models.
\newblock In \emph{International conference on machine learning}, pages 5338--5348. PMLR.

\bibitem[{Krause et~al.(2013)Krause, Stark, Deng, and Fei-Fei}]{car}
Jonathan Krause, Michael Stark, Jia Deng, and Li~Fei-Fei. 2013.
\newblock 3d object representations for fine-grained categorization.
\newblock In \emph{Proceedings of the IEEE international conference on computer vision workshops}, pages 554--561.

\bibitem[{Li et~al.(2023)Li, Li, Savarese, and Hoi}]{blip2}
Junnan Li, Dongxu Li, Silvio Savarese, and Steven Hoi. 2023.
\newblock Blip-2: Bootstrapping language-image pre-training with frozen image encoders and large language models.
\newblock \emph{arXiv preprint arXiv:2301.12597}.

\bibitem[{Liu et~al.(2024)Liu, Roy, Li, Zhong, Sebe, and Ricci}]{fine-grained}
Mingxuan Liu, Subhankar Roy, Wenjing Li, Zhun Zhong, Nicu Sebe, and Elisa Ricci. 2024.
\newblock Democratizing fine-grained visual recognition with large language models.
\newblock \emph{arXiv preprint arXiv:2401.13837}.

\bibitem[{Maji et~al.(2013)Maji, Kannala, Rahtu, Blaschko, and Vedaldi}]{aircraft}
S.~Maji, J.~Kannala, E.~Rahtu, M.~Blaschko, and A.~Vedaldi. 2013.
\newblock \href {https://arxiv.org/abs/1306.5151} {Fine-grained visual classification of aircraft}.
\newblock Technical report.

\bibitem[{Nilsback and Zisserman(2008)}]{flower}
Maria-Elena Nilsback and Andrew Zisserman. 2008.
\newblock Automated flower classification over a large number of classes.
\newblock In \emph{2008 Sixth Indian conference on computer vision, graphics \& image processing}, pages 722--729. IEEE.

\bibitem[{Nishida et~al.(2022)Nishida, Nishida, and Nishioka}]{nishida}
Kosuke Nishida, Kyosuke Nishida, and Shuichi Nishioka. 2022.
\newblock Improving few-shot image classification using machine-and user-generated natural language descriptions.
\newblock \emph{arXiv preprint arXiv:2207.03133}.

\bibitem[{Pratt et~al.(2023)Pratt, Covert, Liu, and Farhadi}]{pratt}
Sarah Pratt, Ian Covert, Rosanne Liu, and Ali Farhadi. 2023.
\newblock What does a platypus look like? generating customized prompts for zero-shot image classification.
\newblock In \emph{Proceedings of the IEEE/CVF International Conference on Computer Vision}, pages 15691--15701.

\bibitem[{Radford et~al.(2021)Radford, Kim, Hallacy, Ramesh, Goh, Agarwal, Sastry, Askell, Mishkin, Clark et~al.}]{clip}
Alec Radford, Jong~Wook Kim, Chris Hallacy, Aditya Ramesh, Gabriel Goh, Sandhini Agarwal, Girish Sastry, Amanda Askell, Pamela Mishkin, Jack Clark, et~al. 2021.
\newblock Learning transferable visual models from natural language supervision.
\newblock In \emph{International conference on machine learning}, pages 8748--8763. PMLR.

\bibitem[{Roth et~al.(2022)Roth, Vinyals, and Akata}]{cbm2}
Karsten Roth, Oriol Vinyals, and Zeynep Akata. 2022.
\newblock Integrating language guidance into vision-based deep metric learning.
\newblock In \emph{Proceedings of the IEEE/CVF Conference on Computer Vision and Pattern Recognition}, pages 16177--16189.

\bibitem[{Rudin(2019)}]{stop}
Cynthia Rudin. 2019.
\newblock Stop explaining black box machine learning models for high stakes decisions and use interpretable models instead.
\newblock \emph{Nature machine intelligence}, 1(5):206--215.

\bibitem[{Shen et~al.(2022)Shen, Li, Hu, Xie, Yang, Zhang, Gan, Wang, Yuan, Liu et~al.}]{cbm3}
Sheng Shen, Chunyuan Li, Xiaowei Hu, Yujia Xie, Jianwei Yang, Pengchuan Zhang, Zhe Gan, Lijuan Wang, Lu~Yuan, Ce~Liu, et~al. 2022.
\newblock K-lite: Learning transferable visual models with external knowledge.
\newblock \emph{Advances in Neural Information Processing Systems}, 35:15558--15573.

\bibitem[{Singh et~al.(2024)Singh, Inala, Galley, Caruana, and Gao}]{interpretablellm}
Chandan Singh, Jeevana~Priya Inala, Michel Galley, Rich Caruana, and Jianfeng Gao. 2024.
\newblock Rethinking interpretability in the era of large language models.
\newblock \emph{arXiv preprint arXiv:2402.01761}.

\bibitem[{Singh et~al.(2022)Singh, Morris, Aneja, Rush, and Gao}]{llm_explain}
Chandan Singh, John~X Morris, Jyoti Aneja, Alexander~M Rush, and Jianfeng Gao. 2022.
\newblock Explaining patterns in data with language models via interpretable autoprompting.
\newblock \emph{arXiv preprint arXiv:2210.01848}.

\bibitem[{Stan et~al.(2024)Stan, Rohekar, Gurwicz, Olson, Bhiwandiwalla, Aflalo, Wu, Duan, Tseng, and Lal}]{lvlm}
Gabriela Ben~Melech Stan, Raanan~Yehezkel Rohekar, Yaniv Gurwicz, Matthew~Lyle Olson, Anahita Bhiwandiwalla, Estelle Aflalo, Chenfei Wu, Nan Duan, Shao-Yen Tseng, and Vasudev Lal. 2024.
\newblock Lvlm-intrepret: An interpretability tool for large vision-language models.
\newblock \emph{arXiv preprint arXiv:2404.03118}.

\bibitem[{Subramanian et~al.(2023)Subramanian, Narasimhan, Khangaonkar, Yang, Nagrani, Schmid, Zeng, Darrell, and Klein}]{modular_code}
Sanjay Subramanian, Medhini Narasimhan, Kushal Khangaonkar, Kevin Yang, Arsha Nagrani, Cordelia Schmid, Andy Zeng, Trevor Darrell, and Dan Klein. 2023.
\newblock Modular visual question answering via code generation.
\newblock \emph{arXiv preprint arXiv:2306.05392}.

\bibitem[{Sur{\'\i}s et~al.(2023)Sur{\'\i}s, Menon, and Vondrick}]{vipergpt}
D{\'\i}dac Sur{\'\i}s, Sachit Menon, and Carl Vondrick. 2023.
\newblock Vipergpt: Visual inference via python execution for reasoning.
\newblock \emph{arXiv preprint arXiv:2303.08128}.

\bibitem[{Wah et~al.(2011)Wah, Branson, Welinder, Perona, and Belongie}]{cub}
C.~Wah, S.~Branson, P.~Welinder, P.~Perona, and S.~Belongie. 2011.
\newblock Caltech-ucsd birds-200-2011 (cub-200-2011).
\newblock Technical Report CNS-TR-2011-001, California Institute of Technology.

\bibitem[{Yang et~al.(2023)Yang, Panagopoulou, Zhou, Jin, Callison-Burch, and Yatskar}]{labo}
Yue Yang, Artemis Panagopoulou, Shenghao Zhou, Daniel Jin, Chris Callison-Burch, and Mark Yatskar. 2023.
\newblock Language in a bottle: Language model guided concept bottlenecks for interpretable image classification.
\newblock In \emph{Proceedings of the IEEE/CVF Conference on Computer Vision and Pattern Recognition}, pages 19187--19197.

\bibitem[{Zhang et~al.(2022)Zhang, Chan, and Mahadevan}]{safety}
Xiaoge Zhang, Felix~TS Chan, and Sankaran Mahadevan. 2022.
\newblock Explainable machine learning in image classification models: An uncertainty quantification perspective.
\newblock \emph{Knowledge-Based Systems}, 243:108418.

\bibitem[{Zhang et~al.(2023)Zhang, Li, Cui, Cai, Liu, Fu, Huang, Zhao, Zhang, Chen et~al.}]{hallucination}
Yue Zhang, Yafu Li, Leyang Cui, Deng Cai, Lemao Liu, Tingchen Fu, Xinting Huang, Enbo Zhao, Yu~Zhang, Yulong Chen, et~al. 2023.
\newblock Siren's song in the ai ocean: a survey on hallucination in large language models.
\newblock \emph{arXiv preprint arXiv:2309.01219}.

\end{thebibliography}

\appendix
\section{Inference Demo}
\label{sec:appendix}
The subsequent three pages feature a demonstration of an inference tool developed using Streamlit, showcasing its application on a randomly selected Dandelion image. The probability assigned by each concept classifier to its prediction is clearly displayed. Additionally, the contribution of each attribute value towards the prediction made by each concept classifier is transparently presented. While it is feasible to render a hierarchical tree in real-time, similar to the format depicted in Figure \ref{fig:snippet1} and Figure \ref{fig:snippet2}, these visualizations are too extensive to be included in print within this thesis. The tool's straightforward visualization and enhanced interpretability facilitate easy debugging and inspection. For example, it is evident from the visualization that the Anther and Petal classifiers, which predicted incorrect labels at a relatively lower probability, highlight potential failure modes and areas for improvement.
\includepdf[pages=-]{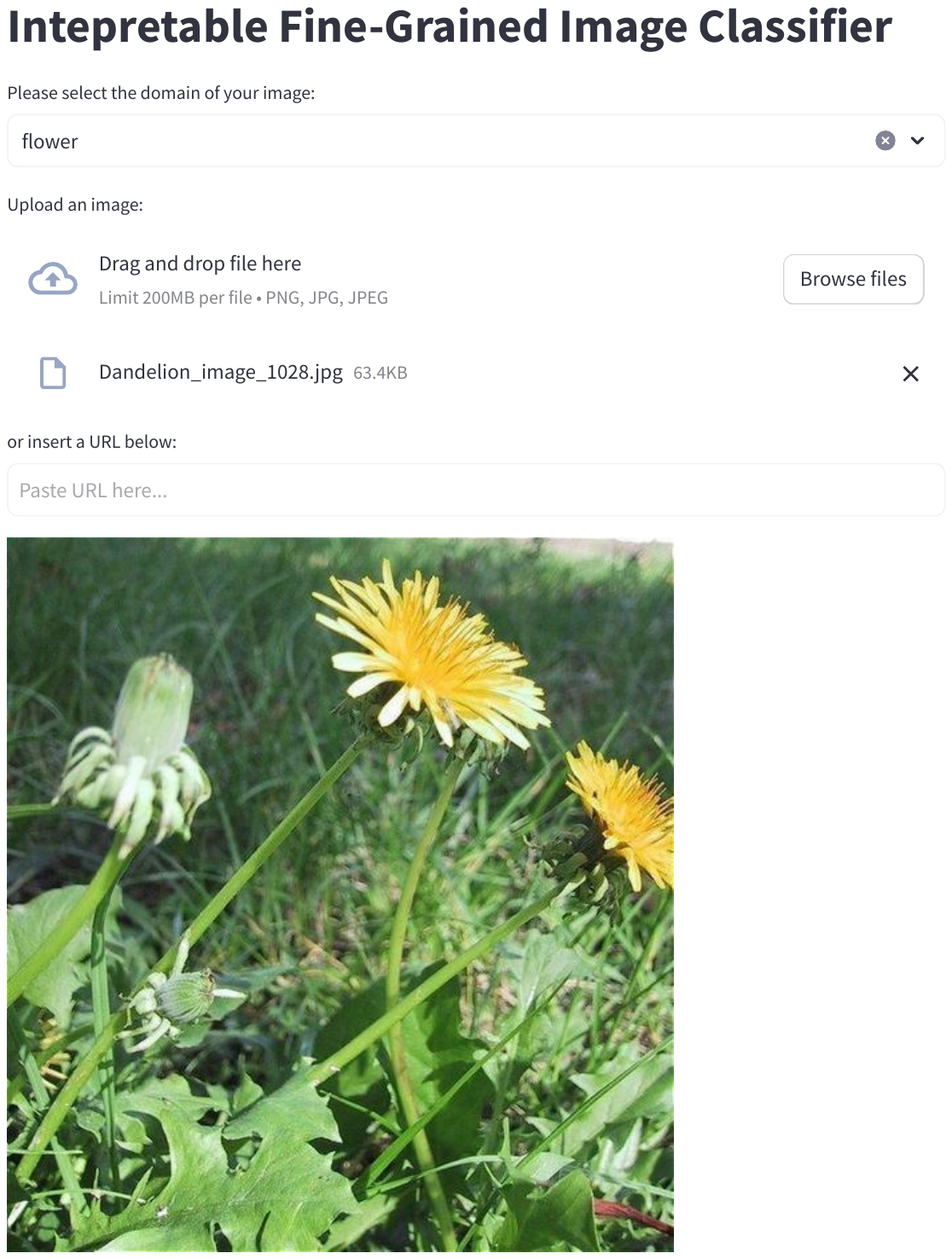}
\end{document}